\pdfoutput=1
\documentclass[10pt,twocolumn,letterpaper]{article}

\usepackage[pagenumbers]{cvpr}      %

\usepackage{microtype}

\newcommand{\bench}{\textup{\textsc{EyeHeight}}\xspace}
\usepackage{multirow}
\usepackage{siunitx}
\usepackage{nicefrac}
\newcommand{\hstar}{h^{\ast}}

\definecolor{cvprblue}{rgb}{0.21,0.49,0.74}
\usepackage[pagebackref,breaklinks,colorlinks,allcolors=cvprblue]{hyperref}

\title{Every Camera Grows Up: Camera-Height Regression in Interactive Video World Models}

\author{Sheng Lu \qquad Guangpin Tao \qquad Tong Lu\textsuperscript{\dag} \qquad Ying Tai\\
State Key Laboratory for Novel Software Technology, Nanjing University\\
{\tt\small \{shenglu,tgpin\}@smail.nju.edu.cn \qquad \{lutong,yingtai\}@nju.edu.cn}
}

\begin{document}
\twocolumn[{%
  \renewcommand\twocolumn[1][]{#1}%
  \maketitle
  \begin{center}
    \includegraphics[width=\textwidth]{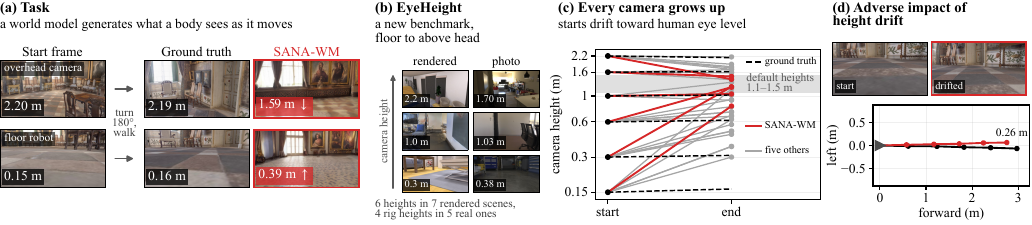}
    \captionof{figure}{\textbf{Camera-height regression in interactive video world models.}
      (a)~From a start frame, a world model generates the views along a camera path, here a \ang{180} turn in place and a
      walk. By frame~160, ground-truth rendering keeps a \SI{2.2}{m} and a \SI{0.15}{m} camera at their heights, whereas
      SANA-WM shows the castle hall from inferred heights of \SI{1.59}{m} and \SI{0.39}{m}.
      (b)~Our new benchmark, \bench, starts cameras at six heights (\SIrange{0.15}{2.2}{m}) in seven rendered scenes and
      at four rig heights in calibrated photographs of five real scenes.
      (c)~On rendered turn rollouts, all six models pull low and high starts toward default heights of
      \SIrange{1.1}{1.5}{m}, near human eye level (medians over start spots).
      (d)~Viewpoint drift alone biases embodied control: re-rendering the castle at the drifted height and pitch of the
      low SANA-WM clip shifts the path endpoint predicted by GNM~\cite{shah2023gnm} by \SI{0.26}{m}, close to its
      \SI{0.28}{m} median shift across turn clips.}
    \label{fig:teaser}
  \end{center}
  \vspace{0.6em}
}]
{\renewcommand{\thefootnote}{\fnsymbol{footnote}}\footnotetext[2]{Corresponding author.}}
\begin{abstract}
  Interactive video world models serve as simulators for embodied agents: from a starting image and an action trajectory, they generate the views the agent would see.
  The starting image fixes how high the camera is, for example \SI{0.15}{m} for a floor robot or \SI{2.2}{m} for an overhead camera, and a faithful simulator must keep this height as the camera moves.
  We show that six recent open-weight models fail to do so: as motion replaces the starting view with generated content, they move the camera toward human eye level, a failure we call \emph{camera-height regression}.
  Our benchmark, \bench, pairs six rendered elevations (\SIrange{0.15}{2.2}{m}) and four calibrated real-world rig heights with scale-free multi-view height estimation.
  All six models lose their initial height during a \ang{180} turn-and-walk rollout: initialized at \SI{0.15}{m}, only 0--20\% of clips end within $1.3\times$ of the starting height as inferred viewpoints converge to human heights of \SIrange{1.1}{1.5}{m}, whereas ground-truth renders remain within a few percent.
  In Matrix-Game models, homography-tracked floor motion stays consistent with constant elevation during rotation even as newly outpainted room geometry yields a higher reconstructed viewpoint.
  Regression grows with rotation angle, begins while the start frame remains in context, and largely reverses in SANA-WM when a \ang{360} turn restores the initial view.
  We trace this pull to human-centric video priors in which floor-level views are nearly absent: in SANA-WM, explicit textual height cues and dense floor-disparity maps do not prevent it, whereas classifier-free text guidance during the first half of denoising strengthens it.
  Disabling early guidance partially recovers height retention without lowering MUSIQ, at a modest cost in caption alignment.
  Combining unguided sampling with height-balanced fine-tuning recovers more retention but costs 5 MUSIQ points, a loss fixed during those same early steps.
  Viewpoint drift alone shifts the paths predicted by visual navigation policies.
\end{abstract}

\section{Introduction}
\label{sec:intro}

Action-conditioned video generators that synthesize future frames from a starting image and a camera trajectory now serve as interactive world models, promoted as simulators in which embodied agents preview the visual consequences of their actions~\cite{bruce2024genie,matrixgame2,worldplay,sanawm}.
The trajectory determines how the camera moves, while the starting image determines where the camera is, including its height above the floor and its tilt.
A robot vacuum at \SI{0.15}{m} and a standing adult at \SI{1.6}{m} see the same room differently, and visuomotor policies act on the view they receive.
To simulate a specific body, a world model must therefore keep the camera height of its starting frame.

This paper shows that six recent open-weight world models do not meet this requirement.
As motion replaces the starting view with generated content, they move the camera toward human eye level, so that low cameras rise and high cameras sink (\cref{fig:teaser}a).
We call this failure \emph{camera-height regression}, measure it across the six models, trace it to its causes, and test remedies.

To measure it, we introduce \bench, which drives the models along constant-height, constant-pitch trajectories from rendered indoor starts (\SIrange{0.15}{2.2}{m}) and calibrated multi-height real photographs (\cref{fig:teaser}b), and recovers camera height from the generated frames with scale-free multi-view 3D reconstruction.
After a \ang{180} turn in place and a forward walk, all six models lose their starting height and drift toward \SIrange{1.1}{1.5}{m} (\cref{fig:teaser}c).
From \SI{0.15}{m}, only 0--20\% of rollouts end within $1.3\times$ of the starting height, whereas ground-truth renders remain within a few percent.
Overhead cameras (\SI{2.2}{m}) sink toward the same band ($\times0.63$--$\times0.85$ in five models), and on real photographs all four models rotating past \ang{100} regress as well.
Comparing 3D reconstruction with floor homographies shows that Matrix-Game turns are geometrically split: local 2D floor motion stays consistent with constant elevation, while newly outpainted walls and furniture imply a higher viewpoint.

\begin{figure*}[t]
  \centering
  \includegraphics[width=\linewidth]{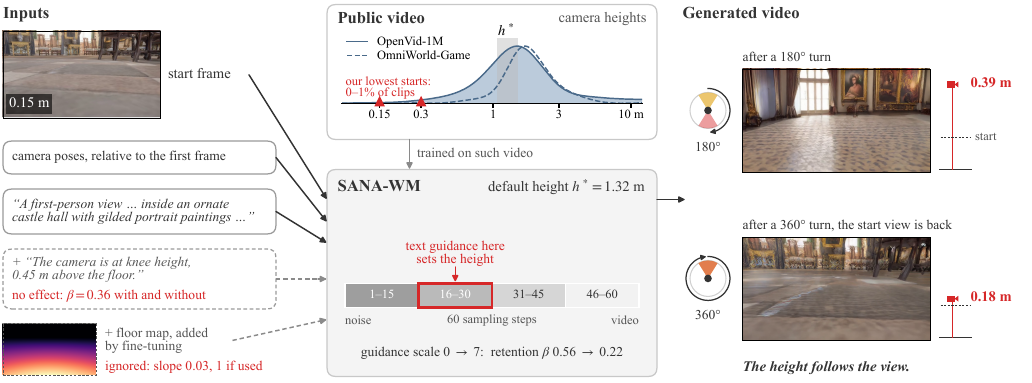}
  \caption{\textbf{Mechanisms of camera-height regression in SANA-WM}, on the castle start of \cref{fig:teaser}.
    \emph{Left:} Supplying metric height via the prompt or a fine-tuned floor-disparity map (dashed) leaves inferred
    height unchanged, and scaling the height encoded in the map yields a response slope of $0.03$ (vs.\ $1$ for faithful
    adherence).
    \emph{Middle:} Public video corpora contain virtually no viewpoints near our lowest starts, and the default heights
    $\hstar$ of the six models (gray) cluster near the human-eye-level mode. Text guidance applied in the
    second quarter of the 60 denoising steps strengthens this pull.
    \emph{Right:} Terminal frames and Pi3-inferred heights after \ang{180} and \ang{360} rotations (dashed: start
    height). Returning to the initial view largely restores the inferred height.
    $\beta$: real-start retention.}
  \label{fig:overview}
\end{figure*}

Controlled experiments trace the regression to scene priors learned in pretraining and strengthened during sampling (\cref{fig:overview}).
Public internet and game videos cluster near \SI{1.6}{m} with floor-level views nearly absent, which pulls both low and high starts toward this mode.
The drift grows with rotation angle and begins while the start frame is still in context, so keeping that frame does not prevent it.
In SANA-WM, the drift largely reverses when a \ang{360} turn brings the starting view back.
Supplying metric height to SANA-WM does not stop the drift either.
Textual height cues have no effect, and a fine-tuned floor-disparity branch responds to counterfactual height scaling with a slope of only $0.03$, where faithful use would give $1$.
Stronger classifier-free text guidance deepens the regression, and most of this effect comes from the first half of the denoising steps.

The drift matters for embodied control (\cref{fig:teaser}d): counterfactually re-rendering ground-truth scenes at each rollout's drifted height and pitch shifts the path endpoints predicted by seven navigation policies by \SIrange{0.12}{1.18}{m}, more than mild blur does on five of them.
Among remedies for SANA-WM, unguided sampling lifts real-start retention from $\beta = 0.36$ to $0.56$ at unchanged MUSIQ and slightly weaker caption alignment.
Adding height-balanced LoRA fine-tuning reaches $\beta = 0.68$, although rollouts still drift toward a lower default height.
It costs 5 MUSIQ points, a loss that arises in the same early steps that set height, and post-training RL adds little.

Our contributions are as follows.
\begin{itemize}
  \item We introduce \bench, a benchmark for camera-height preservation with rendered starts from \SIrange{0.15}{2.2}{m} and calibrated real starts, together with a scale-free height measurement validated on constant and drifting renders and cross-checked with VGGT, monocular depth, and floor homographies.
  \item We show that all six open world models regress toward human eye level from both low and high starts, that Matrix-Game turns separate floor motion from scene scale, and that the drift alone shifts the predictions of seven navigation policies.
  \item We link the regression to human-height video priors, show that neither keeping the start frame in context nor supplying height inputs prevents it, and identify early-step text guidance as a factor that strengthens it.
  \item We evaluate fine-tuning, reinforcement learning, and guidance schedules on SANA-WM and find that height retention and perceptual quality are coupled in the early denoising steps.
\end{itemize}
We will publicly release \bench and the code.

\section{Related Work}
\label{sec:related}

\paragraph{Interactive video world models.}
Action-conditioned video generators have evolved from closed-domain games~\cite{bruce2024genie,valevski2025gamengen,
alonso2024diamond} to open-domain scenes synthesized from a single image~\cite{matrixgame2,matrixgame3,yume15,
worldplay,lingbotworld,sanawm}, alongside hybrid pipelines that render explicit 3D caches along camera
paths~\cite{ren2025gen3c,huang2025voyager}.
Existing benchmarks score visual quality, action adherence, and 3D consistency~\cite{huang2024vbench,
duan2025worldscore,worldmark}, test generalization to unseen robot bodies visible in the frame~\cite{xeworld}, or fix
one non-human vantage point such as UAV video~\cite{motionscape}.
\bench complements them by sweeping the starting camera height to test whether a model keeps the sensor elevation of
the body that carries the camera.

\paragraph{Camera control and viewpoint priors.}
Pose-conditioned video diffusion typically encodes trajectories as relative Pl\"ucker rays~\cite{wang2024motionctrl,
he2025cameractrl,bahmani2025ac3d}, which fix inter-frame motion but leave floor elevation and gravity tilt implicit in
perspective.
Canonical-camera biases also affect object novel-view synthesis~\cite{liu2023zero123} under non-standard elevations or
focal lengths~\cite{li2024era3d}, and errors under out-of-distribution scene motion have motivated explicit geometric
conditioning~\cite{geostream}, unified lens encodings~\cite{crepe}, and gravity-grounded world
states~\cite{puffinworld}.
We show that camera height regresses toward human eye level even when relative rotation is followed, and that supplying
explicit ground elevation does not prevent it.

\paragraph{Camera height in embodied perception.}
Perception and navigation degrade across camera heights~\cite{kumar2025charm3r,gvnav}, which has motivated
multi-embodiment navigation policies~\cite{shah2023gnm,shah2023vint,sridhar2024nomad} and morphology-specific video
generators~\cite{quadreamer}.
We show that height drift in general-purpose world models biases such policies.

\section{Benchmarking Camera-Height Preservation}
\label{sec:benchmark}

An interactive world model takes an initial image $I_0$, an optional prompt, and relative camera poses or actions to
synthesize future frames.
Geometrically, $I_0$ fixes the agent's embodiment: its height $h_0$ above the floor and its pitch.
Every trajectory in \bench restricts motion to horizontal translation and in-place yaw, holding true height and pitch
constant, and \bench measures whether rollouts preserve $h_0$ or drift.

\begin{figure*}[t]
  \centering
  \includegraphics[width=\linewidth]{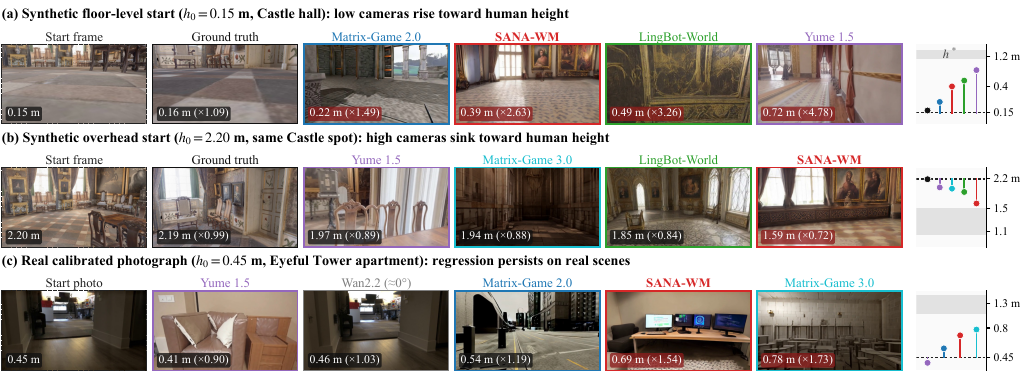}
  \caption{\textbf{Qualitative rollouts after a \ang{180} turn and forward walk (frame 160).}
    \textbf{(a)}~Initialized near the floor (\SI{0.15}{m}) in the synthetic castle hall, ground-truth rendering keeps
    the low perspective, whereas the world models raise the camera toward human eye height.
    \textbf{(b)}~Initialized near the ceiling (\SI{2.20}{m}) at the same spot, the models pull the viewpoint down
    toward the same band $\hstar \in [\SI{1.06}{m}, \SI{1.51}{m}]$ (right gauges).
    \textbf{(c)}~On a calibrated real photograph (\SI{0.45}{m}, Eyeful Tower), models that execute the turn also raise
    the viewpoint to seated or standing human perspectives.}
  \label{fig:qual_main}
\end{figure*}

\paragraph{Camera trajectories.}
Each script spans 160 frames beyond $I_0$ with zero vertical motion.
\emph{Forward} translates ahead at \SI{1.5}{cm}/frame (\SI{2.4}{m} total).
\emph{Turn} rotates rightward in place at \ang{2.8}/frame for 64 frames (\ang{180}) before walking forward for 96
frames (\SI{1.44}{m}): the rotation forces outpainting of the unseen room behind the camera, while the walk adds
parallax to probe 3D consistency with $h_0$.
To study rotation angle and context, we add turns of \ang{45}, \ang{90}, \ang{135}, and \ang{360} at \ang{2.8}/frame
(each followed by the same walk), a half-rate slow turn (\ang{180} over 128 frames, then 32 forward), and a static
script.

\paragraph{Synthetic and real starting views.}
With Habitat-Sim~\cite{savva2019habitat,szot2021habitat}, we render synthetic starts across seven indoor scenes: five
ReplicaCAD apartments with baked lighting~\cite{szot2021habitat}, a Matterport3D house~\cite{chang2017matterport3d},
and a Habitat castle hall.
At three open spots per scene, we mount cameras at six embodiment heights: \SI{0.15}{m} (robot vacuum), \SI{0.3}{m}
(pet/small robot), \SI{0.6}{m}, \SI{1.0}{m} (child/wheeled robot), \SI{1.6}{m} (standing adult), and \SI{2.2}{m}
(mounted/aerial sensor).
Two spots lack clearance at \SI{2.2}{m}, leaving 124 pinhole starts ($1280\times704$, \ang{70} horizontal FoV, \ang{10}
downward pitch, zero roll), paired with ground-truth renders for all 124 forward starts and 59 turn starts (ten spots).
To evaluate photographs free of rendering gaps, we curate 52 real starts from Eyeful Tower~\cite{xu2023vrnerf}, taking
13 spots across five scenes at four vertical rig tiers (\SIrange{0.36}{0.46}{m}, \SIrange{0.63}{0.92}{m},
\SIrange{1.01}{1.32}{m}, and \SIrange{1.64}{1.72}{m}) and reprojecting each photograph via dataset calibration into our
synthetic pinhole camera so that starts at a spot differ only in height.

\paragraph{Prompts and models.}
Prompted models share one elevation-neutral room caption per scene across all heights.
We prepend ``A first-person view'' for LingBot-World and Matrix-Game 3.0, SANA-WM's standard prefix ``A
first-person view from a strictly stationary observation point,'' and ``First-person view'' plus a one-sentence motion
summary for Wan2.2.
Yume 1.5 receives the bare caption and appends motion tokens, while Matrix-Game 2.0 takes no text.
Ablating ``first-person'' in SANA-WM leaves real-start retention unchanged with guidance ($\beta = 0.32$ vs.\ $0.36$)
and without ($0.54$ vs.\ $0.56$, supplementary material).

We benchmark six open models of different sizes, architectures, and control types.
SANA-WM~\cite{sanawm} (\num{2.6}B, bidirectional, base first stage) and LingBot-World~\cite{lingbotworld} (v2,
\num{1.3}B, causal) take camera poses and intrinsics.
Matrix-Game 2.0~\cite{matrixgame2} and 3.0~\cite{matrixgame3} take keyboard (`W`) and mouse-yaw inputs (calibrated on
2.0 to reach \ang{180}), Yume 1.5~\cite{yume15} takes movement and rotation phrases per 32-frame chunk, and
Wan2.2-TI2V-5B~\cite{wan2025} takes prompt-only motion.
All run at native resolution with default sampling (30 of 50 steps for Wan2.2) and true intrinsics where accepted.

\paragraph{Scale-free multi-view height measurement and metrics.}
Because monocular metric depth priors pull extreme views toward human height, we measure \emph{within-clip height
ratios} from a unified multi-view reconstruction: Pi3~\cite{wang2025pi3} reconstructs every fourth frame in one pass
under a shared global scale while reading executed yaw.
Segmenting floor and rug pixels via ADE20K-trained SegFormer-B5~\cite{xie2021segformer,zhou2017ade20k} and fitting a
plane to the lowest dense horizontal ground layer gives the \emph{reconstruction-inferred height} $h(t)$.
The terminal ratio $r = h(T)/h(0)$ ($h(T)$: median over the final 20\% of frames) cancels scene scale, giving
$h_\mathrm{end} = h_0\, r$.
Clips lacking a reconstructable scene (no initial floor) count as failing to keep height, while continuous fits use
measurable clips.
We validate this measurement on ground-truth renders and against three other estimators (supplementary material).
On constant-height renders, Pi3 recovers $\beta = 0.95$ on the turn script and $0.99$ on the forward script, and it
follows synthetic $2\times$ and $0.6\times$ height ramps during the turn to within 12\%.
VGGT~\cite{wang2025vggt} reads the same constant renders at $\beta = 1.03$ and $1.01$ and confirms the regression of
every model ($\beta = 0.15$--$0.51$, against $0.18$--$0.67$ with Pi3).
MoGe-2~\cite{wang2025moge2} estimates the calibrated heights of real starts with median errors of \SIrange{3}{9}{cm}.
Tracked floor homographies, which measure how much height change the floor motion alone supports, match constant
renders within 3\% and drifting renders within 9\%.

Per clip we report $|\log r|$ and whether the clip keeps its height ($|\log r| < \log 1.3$).
Per model and script we fit
\begin{equation}
  \log h_\mathrm{end} = \beta \log h_0 + (1 - \beta) \log \hstar,
  \label{eq:fit}
\end{equation}
or $h_\mathrm{end} = \hstar (h_0 / \hstar)^{\beta}$, by least squares, which gives the \emph{retention}
$\beta \in [0, 1]$ ($1$: ideal, $0$: collapse) and the \emph{default height} $\hstar$ (undefined near $\beta = 1$).
For example, with $\beta = 0.5$ and $\hstar = \SI{1.3}{m}$, a \SI{0.15}{m} start would end at \SI{0.44}{m} and a
\SI{2.2}{m} start at \SI{1.69}{m}.
Tables report 95\% intervals from 1,000 bootstrap resamples over start spots.
Scene-level intervals and paired comparisons between settings are given in the supplementary material.

\section{Cameras Regress Toward a Default Height}
\label{sec:regression}

\begin{table*}[t]
  \centering
  \footnotesize
  \setlength{\tabcolsep}{2.8pt}
  \begin{tabular}{@{}l l r ccrrr c r cr@{}}
    \toprule
    & & & \multicolumn{6}{c}{Synthetic starts, turn} & Forward & \multicolumn{2}{c}{Real starts, turn} \\
    \cmidrule(lr){4-9} \cmidrule(lr){10-10} \cmidrule(l){11-12}
    Model & Control & Yaw & $\beta$ & $\hstar$ & $r$ at \SI{0.15}{m} & Kept & $r$ at $\approx$\ang{90} & $r$ at \SI{2.2}{m} & $\beta$ & $\beta$ & $r$, lowest \\
    \midrule
    Ground truth & -- & \ang{180}--\ang{182} & 0.95 {\scriptsize[0.90, 0.99]} & -- & $\times$1.05 & 10/10 & $\times$1.01 & $\times$0.97 & 0.99 & -- & -- \\
    \midrule
    SANA-WM~\cite{sanawm} & poses & \ang{167}--\ang{209} & 0.18 {\scriptsize[0.04, 0.32]} & 1.32 & $\times$5.47 & 0/21 & $\times$4.49 {\scriptsize(\ang{90})} & $\times$0.63 & 0.99 & 0.36 {\scriptsize[0.27, 0.46]} & $\times$1.85 \\
    Matrix-Game 2.0~\cite{matrixgame2} & keys, mouse & \ang{171}--\ang{174} & 0.64 {\scriptsize[0.54, 0.75]} & 1.51 & $\times$2.47 & 2/18 & $\times$1.02 {\scriptsize(\ang{75})} & $\times$0.81 & 0.83 & 0.62 {\scriptsize[0.45, 0.78]} & $\times$1.32 \\
    Matrix-Game 3.0~\cite{matrixgame3} & keys, mouse & \ang{176}--\ang{205} & 0.67 {\scriptsize[0.60, 0.73]} & 1.14 & $\times$2.03 & 4/20 & $\times$1.07 {\scriptsize(\ang{92})} & $\times$0.85 & 0.66 & 0.79 {\scriptsize[0.70, 0.88]} & $\times$1.15 \\
    LingBot-World~\cite{lingbotworld} & poses & \ang{97}--\ang{111} & 0.64 {\scriptsize[0.55, 0.74]} & 1.14 & $\times$2.47 & 2/21 & $\times$1.58 {\scriptsize(\ang{94})} & $\times$0.77 & 0.48 & 0.56 {\scriptsize[0.45, 0.69]} & $\times$1.43 \\
    Yume 1.5~\cite{yume15} & text actions & \ang{93}--\ang{119} & 0.44 {\scriptsize[0.33, 0.55]} & 1.06 & $\times$4.22 & 1/20 & $\times$2.94 {\scriptsize(\ang{92})} & $\times$0.79 & 0.40 & \textcolor{gray}{0.80 {\scriptsize[0.54, 1.03]}} & \textcolor{gray}{$\times$0.90} \\
    Wan2.2-TI2V-5B~\cite{wan2025} & prompt & $\approx$\ang{0} & 0.33 {\scriptsize[0.22, 0.44]} & 1.45 & $\times$7.67 & 0/21 & -- & $\times$0.98 & 0.78 & \textcolor{gray}{1.00 {\scriptsize[0.99, 1.01]}} & \textcolor{gray}{$\times$0.99} \\
    \bottomrule
  \end{tabular}
  \caption{\textbf{Reconstruction-inferred camera height after a \ang{180} turn-and-walk rollout.}
    \emph{Yaw:} Pi3-measured rotation for a commanded \ang{180} turn (range of medians across the six start heights).
    $\beta$, $\hstar$~(m): fit of \cref{eq:fit} with 95\% spot-bootstrap intervals.
    $r$: median end-to-start height ratio (frames 128--160, post-turn walk) from \SI{0.15}{m} and \SI{2.2}{m}.
    \emph{Kept:} \SI{0.15}{m} clips ending within $1.3\times$ of $h_0$, excluding only clips with a valid initial floor
    but none at the end (clips lacking an initial floor count as not kept).
    $r$ at $\approx$\ang{90}: ratio from \SI{0.15}{m} at the end of the turn whose executed yaw (in parentheses) lies
    closest to \ang{90}, cut there (ground truth from the \ang{180} turn).
    \emph{Forward:} $\beta$ on pure forward walks.
    \emph{Real starts:} Eyeful Tower photos (lowest tier \SIrange{0.36}{0.46}{m}), where gray denotes models moving
    little on real photos (Yume 1.5: \ang{77}, Wan2.2: stationary).
    SANA-WM at \SI{0.15}{m} reports $r$ over 4/21 clips: in the other 17, Pi3 finds no floor in frame~0, with no hard
    cut in frame differences during the turn (VGGT reads $\times5.88$ across all 21 and MoGe-2 reads $\times7.30$, see
    supplementary material).}
  \label{tab:main}
\end{table*}

\paragraph{Low and high starts converge to a default height.}
Inferred camera heights move toward human eye level (\cref{fig:qual_main}a,b, \cref{tab:main}).
After the \ang{180} turn-and-walk script, retention is $\beta = 0.18$--$0.67$ for the six models against $0.95$ for
ground-truth renders, and the fitted default heights lie between \SI{1.06}{m} and \SI{1.51}{m}.
Low starts suffer most: from \SI{0.15}{m}, only 0--20\% of rollouts end within $1.3\times$ of $h_0$.
The pull acts in both directions.
Among the five models that rotate, \SI{0.15}{m} cameras rise by $\times2.0$--$\times5.5$ and \SI{2.2}{m} cameras
sink to $\times0.63$--$\times0.85$.
The downward factor is smaller because \SI{2.2}{m} lies closer to $\hstar$ in log-height.
Wan2.2 barely rotates under prompt-only control, yet it still lifts \SI{0.15}{m} cameras by $\times7.7$ and leaves
\SI{2.2}{m} starts nearly unchanged.
Pitch drifts along with height: Matrix-Game 2.0 and 3.0 level \SI{2.2}{m} cameras started at \ang{-10} to within
\ang{1} of horizontal (supplementary material).
Real photographs show the same pattern (\cref{fig:qual_main}c, \cref{tab:main} right), and the four models that turn
by more than \ang{100} all regress, with $\beta$ ranging from $0.36$ for SANA-WM to $0.79$ for Matrix-Game 3.0.

\begin{figure}[t]
  \centering
  \includegraphics[width=\linewidth]{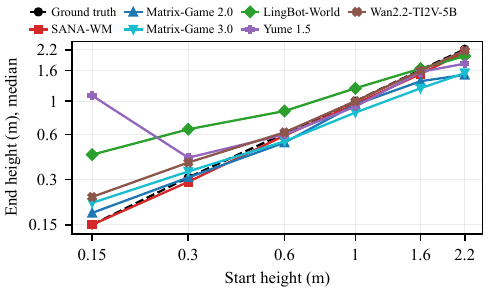}
  \caption{\textbf{End height against start height on pure forward walks} (synthetic starts, medians of $h_0\, r$ over
    19--21 spots).
    A model that keeps its height lies on the dashed ground truth, and \cref{fig:teaser}c plots the turn script.}
  \label{fig:end_height}
\end{figure}

\paragraph{SANA-WM drifts only while outpainting, the others also while walking.}
Comparing pure forward walks (\cref{tab:main}, Forward, and \cref{fig:end_height}) with separate reconstructions of the
turn and of the post-turn walk shows three patterns.
SANA-WM drifts only while it outpaints during the rotation ($\times3.95$ from \SI{0.15}{m}) and holds height whenever
visible structure anchors the scene, on forward walks ($\beta = 0.99$) as well as after the turn.
LingBot-World and Yume 1.5, which generate causally or chunk by chunk, climb whenever they synthesize unseen geometry,
during turns and also on forward walks ($\times2.95$ and $\times7.28$ from \SI{0.15}{m} over the forward script).
Matrix-Game 2.0 and 3.0 pull high cameras down on forward walks, with default heights of \SI{0.42}{m} and
\SI{0.54}{m}.
On generated clips that drift, the turn and walk ratios measured separately multiply to less than the full-clip ratio
(supplementary material).

\begin{figure}[t]
  \centering
  \includegraphics[width=\linewidth]{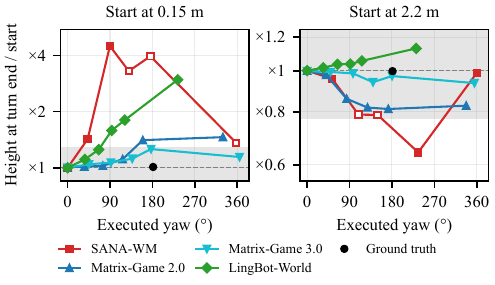}
  \caption{\textbf{Elevation drift against executed rotation.}
    End-to-start height ratios after turns of \ang{45}, \ang{90}, \ang{135}, \ang{180}, and \ang{360} at
    \ang{2.8}/frame (each clip cut at turn end and reconstructed alone), plus the static baseline (\ang{0}).
    Markers plot medians over measurable clips from 21 starts at \SI{0.15}{m} and 19 at \SI{2.2}{m} (open: fewer than
    half measurable) at Pi3's measured yaw.
    Gray band: within $1.3\times$ of $h_0$.}
  \label{fig:dose}
\end{figure}

\paragraph{Drift accumulates with turn angle and reverses only in SANA-WM.}
Without motion commands, every model except Yume 1.5 holds $h_0$ to within 1\%.
Cutting rollouts at the end of turns of increasing angle shows the drift accumulating as rotation replaces $I_0$
(\cref{fig:dose}).
From \SI{0.15}{m}, Matrix-Game 2.0/3.0 and LingBot-World rise by at most $\times1.11$ after \ang{45} and by up to
$\times1.80$ after \ang{180}, while SANA-WM already reaches $\times4.49$ after \ang{90}.
Completing a \ang{360} turn separates streaming from bidirectional generation.
Matrix-Game 2.0/3.0 stay near their mid-turn heights and LingBot-World, which stalls at \ang{234}, rises further, whereas
SANA-WM largely returns toward $h_0$ once $I_0$ comes back into view ($\times1.36$ after \ang{360} against $\times3.95$
after \ang{180}, from \SI{0.15}{m}).
Two effects likely contribute: SANA-WM ties scene scale to $I_0$ whenever $I_0$ is visible, and the reconstruction
aligns scale more easily when the loop closes.

\begin{figure}[t]
  \centering
  \includegraphics[width=\linewidth]{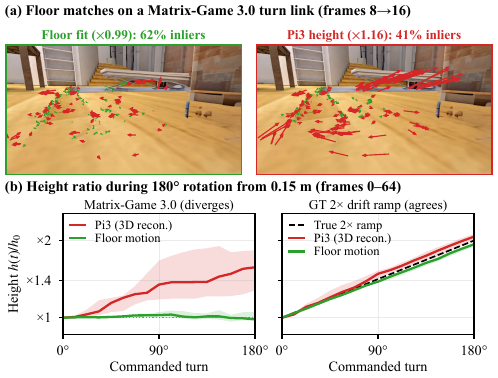}
  \caption{\textbf{Geometric split between local floor motion and 3D scene scale.}
    \textbf{(a)}~On an 8-frame turn link of Matrix-Game 3.0 from \SI{0.15}{m}, a constant-height floor homography
    ($\times0.99$) maps 62\% of tracked ground corners within $2$\,px (green vectors, residual $\times 8$), whereas
    forcing Pi3's inferred rise ($\times1.16$) drops inliers to 41\% (red errors).
    \textbf{(b)}~Across the 64-frame rotation (medians and IQRs), tracked floor motion stays flat while Pi3 climbs on
    Matrix-Game 3.0 (\emph{left}), whereas both rise together on a physical $2\times$ ground-truth ramp (\emph{right}).}
  \label{fig:floor_split}
\end{figure}

\paragraph{Local floor flow and scene scale decouple during turns.}
When the executed yaw is matched at \ang{80}--\ang{100}, Pi3 already infers clear rises from \SI{0.15}{m} for
Matrix-Game 2.0/3.0, LingBot-World, and Yume 1.5 ($\times1.34$--$\times4.36$, against $\times1.03$ for ground truth,
supplementary material).
For Matrix-Game, the tracked floor disagrees with this rise (\cref{fig:floor_split}).
On the 16 Matrix-Game 2.0 and 19 Matrix-Game 3.0 clips that floor tracking covers, the floor motion supports constant
height ($\times1.00$ and $\times1.02$), while Pi3 infers $\times1.31$ and $\times1.38$ over the same clips.
Matrix-Game turns therefore produce geometrically split scenes: the floor texture rotates at constant elevation while
newly outpainted room structure implies a higher camera.

\paragraph{Viewpoint drift biases visual navigation.}
To isolate the camera pose, we re-render the true 3D scene along each rollout's executed yaw twice, at the start height
and pitch and at the height and pitch that Pi3 measures in the rollout (\cref{fig:teaser}d).
On both renders we run seven navigation policies every 8 walk frames of both scripts: GNM~\cite{shah2023gnm},
ViNT~\cite{shah2023vint}, NoMaD~\cite{sridhar2024nomad} in exploration and image-goal modes,
MBRA~\cite{hirose2025mbra}, S2E~\cite{s2e}, and FlowPilot~\cite{flowpilot}.
The pose drift alone moves the predicted path endpoint by \SIrange{0.12}{1.18}{m}, depending on policy and world model.
This is $1.6$ to $6.2$ times our measurement-noise floor and 29--64\% of the shift caused by the generated video
itself.
For GNM, ViNT, both NoMaD modes, and MBRA, the shift exceeds that of a mild blur (half resolution, Gaussian
$\sigma = 1.5$\,px, supplementary material).

\section{Why Cameras Regress}
\label{sec:mechanism}

\paragraph{Public video concentrates near human eye level.}
We measure camera heights in two public corpora (\cref{fig:overview}, middle): with MoGe-2 on 339 of 1,500 random
OpenVid-1M~\cite{nan2024openvid} clips that show the floor, and with ground-truth depth on 161 of 196
OmniWorld-Game~\cite{zhou2025omniworld} scenes.
On these scenes, MoGe-2 heights match the ground truth at a median ratio of 1.01.
Both corpora concentrate near standing adult eye level, with medians of \SI{1.56}{m} and \SI{1.72}{m}.
Views near our lowest starts are almost absent (0--1\% of clips within 20\% of \SI{0.15}{m} or \SI{0.3}{m}), whereas
20--34\% of OpenVid-1M clips lie within 20\% of the six default heights $\hstar$ (\cref{tab:main}).
Elevated views are common (17\% and 26\% of clips within 20\% of \SI{2.2}{m}), yet \SI{2.2}{m} starts still sink
toward $\hstar$.
The regression therefore reflects a pull toward the most frequent training heights, since it also affects viewpoints
that are common in the training data.

\paragraph{Context retention does not anchor height.}
If the models drifted because they forget $I_0$, the drift would start only after $I_0$ leaves their context, but it
starts while $I_0$ is still there.
LingBot-World keeps its first six latent frames as an attention sink, yet from \SI{0.15}{m} it rises by $\times1.58$
by the end of a commanded \ang{135} turn (\cref{fig:dose}), and SANA-WM, which attends to all frames, reaches $\times4.49$ after
\ang{90}.
Halving the turn speed of Matrix-Game 2.0/3.0 keeps $I_0$ in view twice as long, but the cameras still rise by
$\times1.26$ and $\times1.23$ by the end of the turn.

\paragraph{Explicit textual and geometric height cues are ignored.}
SANA-WM receives relative poses without metric height, so we also supply the height explicitly.
Appending it to the prompt (``The camera is at knee height, 0.45 m above the floor.'') leaves real-start retention
unchanged at $\beta = 0.36$.
We then fine-tune a zero-initialized Pl\"ucker branch that encodes per-pixel floor disparity (the sine of the downward
ray angle divided by the camera height for pixels below the horizon, and zero above it).
The model largely ignores this input.
When we halve or double the encoded height, the inferred height barely responds, with a log-log slope of $0.03$ under
default guidance and $0.06$ without guidance, where faithful use would give $1$.
Real-start retention is also no better than with plain fine-tuning.

\begin{figure}[t]
  \centering
  \includegraphics[width=\linewidth]{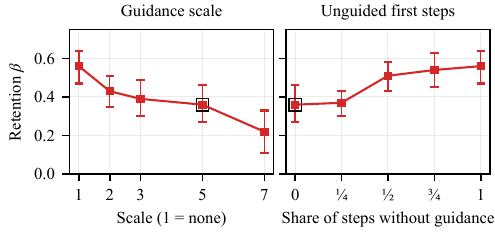}
  \caption{\textbf{Classifier-free text guidance vs.\ height retention in SANA-WM (real starts).}
    Retention $\beta$ across guidance scales (\emph{left}) and across the share of the 60 denoising steps run unguided
    before scale-5 guidance begins (\emph{right}), with 52 turn clips per point and 95\% intervals over 13 rig spots.
    Framed: default.}
  \label{fig:guidance}
\end{figure}

\paragraph{Early-step text guidance strengthens the pull.}
SANA-WM applies classifier-free guidance (CFG)~\cite{ho2022cfg} to the text condition only, at scale 5 and with
identical camera poses in both branches.
Since captioned pretraining videos mostly show eye-level views, text guidance pushes samples toward human viewpoints.
On real starts, retention falls steadily as the scale grows, from $\beta = 0.56$ without guidance to $0.22$ at scale 7
(\cref{fig:guidance}, left).
Turning guidance off keeps MUSIQ~\cite{ke2021musiq} and VBench~\cite{huang2024vbench} consistency at the default level
and costs some caption alignment (CLIP~\cite{radford2021clip} similarity $0.193$ against $0.221$).
Delaying the start of guidance shows when the pull acts (\cref{fig:guidance}, right).
Leaving the first quarter of the 60 steps unguided changes nothing ($\beta = 0.37$), while leaving the first
half~\cite{kynkaanniemi2024interval} unguided recovers most of the height ($\beta = 0.51$) and keeps the global layout
of the unguided rollout for each seed.
Guidance therefore acts on camera height mainly during the second quarter of denoising.
Its effect is smaller on synthetic starts ($\beta = 0.29$ without guidance and $0.18$ with it).
Across schedules, MUSIQ does not follow retention.
It is highest at scales 2--3 and with the first quarter unguided, where retention stays moderate, while the settings
with the highest retention match the default quality.

\section{Mitigations and Quality Trade-offs}
\label{sec:remedies}

\begin{figure*}[t]
  \centering
  \includegraphics[width=\linewidth]{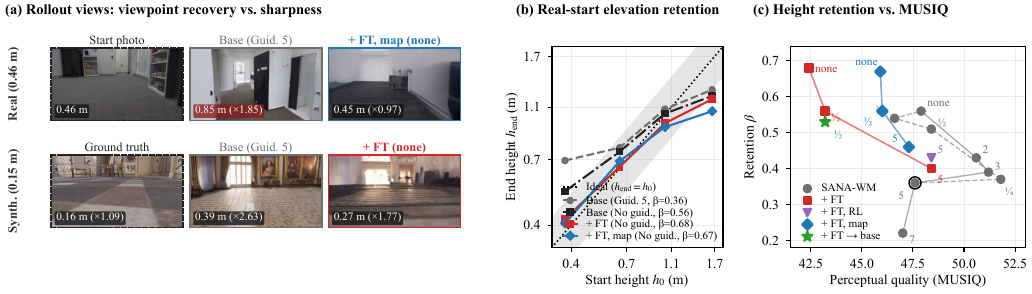}
  \caption{\textbf{Mitigating camera-height regression and the early-denoising quality trade-off in SANA-WM.}
    \textbf{(a)}~Terminal views (frame 160) from a real Eyeful Tower start (\SI{0.46}{m}) and synthetic castle start
    (\SI{0.15}{m}): pairing height-balanced fine-tuning with unguided sampling restores low viewpoints while visibly
    softening high-frequency detail.
    \textbf{(b)}~Median terminal height across the four real-start rig tiers (52 clips, gray band within $1.3\times$ of
    $h_0$).
    \textbf{(c)}~Retention $\beta$ vs.\ perceptual quality (MUSIQ) across all 16 guidance scales, step intervals,
    fine-tuning (FT), RL, floor-map, and base-relay settings on real starts (circled: default).}
  \label{fig:remedies}
\end{figure*}

\Cref{sec:mechanism} points to two causes that can be acted on: low viewpoints are nearly absent from training videos,
and early text guidance strengthens the pull toward eye level.
We therefore evaluate three remedies on SANA-WM (\cref{tab:remedies}, \cref{fig:remedies}): height-balanced fine-tuning
to supply missing low-elevation views, online reinforcement learning rewarding multi-view height retention, and
early-step guidance scheduling.
Combining fine-tuning with unguided sampling preserves the most height but lowers perceptual quality, a loss that arises
in the first half of denoising, where camera height is also set.

\begin{table}[t]
  \centering
  \footnotesize
  \setlength{\tabcolsep}{2.6pt}
  \begin{tabular}{@{}l l cccr c@{}}
    \toprule
    & & \multicolumn{4}{c}{Real starts, turn} & Synth. \\
    \cmidrule(lr){3-6} \cmidrule(l){7-7}
    Model & Guid. & $\beta$ & $r_\mathrm{low}$ & Kept & MUSIQ & $\beta$ \\
    \midrule
    SANA-WM & 5 & 0.36 {\scriptsize[0.27, 0.46]} & $\times$1.85 & 48\% & 47.6 & 0.18 \\
            & none & 0.56 {\scriptsize[0.47, 0.64]} & $\times$1.42 & 60\% & 47.9 & 0.29 \\
            & late & 0.51 {\scriptsize[0.43, 0.58]} & $\times$1.63 & 60\% & 48.4 & 0.23 \\
    \midrule
    + FT & 5 & 0.40 {\scriptsize[0.31, 0.51]} & $\times$1.54 & 54\% & 48.4 & 0.35 \\
            & none & 0.68 {\scriptsize[0.58, 0.77]} & $\times$1.12 & 79\% & 42.4 & 0.56 \\
            & late & 0.56 {\scriptsize[0.45, 0.67]} & $\times$1.15 & 71\% & 43.2 & -- \\
    + FT, RL & 5 & 0.43 {\scriptsize[0.32, 0.53]} & $\times$1.43 & 56\% & 48.4 & 0.38 \\
    \midrule
    + FT, map & 5 & 0.46 {\scriptsize[0.35, 0.58]} & $\times$1.37 & 60\% & 47.3 & 0.34 \\
            & none & 0.67 {\scriptsize[0.63, 0.73]} & $\times$1.08 & 77\% & 45.9 & 0.59 \\
            & late & 0.56 {\scriptsize[0.48, 0.64]} & $\times$1.30 & 69\% & 46.0 & -- \\
    \bottomrule
  \end{tabular}
  \caption{\textbf{Mitigating camera-height regression in SANA-WM.}
    \emph{FT:} height-balanced LoRA fine-tuning, \emph{RL:} subsequent reinforcement learning, \emph{map:} fine-tuning
    with the floor-disparity map (\cref{sec:mechanism}).
    \emph{Guid.:} scale-5 text guidance across all steps, disabled (\emph{none}), or restricted to the second half
    (\emph{late}).
    \emph{Real starts:} $\beta$ with 95\% intervals, median ratio $r_\mathrm{low}$ from the lowest rig tier
    (\SIrange{0.36}{0.46}{m}), share kept within $1.3\times$ of $h_0$, and MUSIQ averaged over frames 40, 80, 120, and
    160.
    \emph{Synth.:} synthetic-start $\beta$.}
  \label{tab:remedies}
\end{table}

\paragraph{With guidance, fine-tuning helps mainly on synthetic starts and RL adds little.}
We render 81-frame ground-truth turns from 258 new starts in our seven synthetic scenes (42--44 per height, at least
\SI{1.5}{m} from benchmark spots, two per height held out for validation).
On these clips we train a rank-32 LoRA on SANA-WM's self-attention, cross-attention, and camera layers with its
unguided flow-matching loss and keep epoch 1 (64 steps).
On synthetic starts, fine-tuning doubles retention from $\beta = 0.18$ to $0.35$.
It also lowers the default height from \SI{1.32}{m} to \SI{0.83}{m}, so \SI{2.2}{m} starts sink further, to
$\times0.50$.
Fine-tuning on eye-level (\SI{1.6}{m}) views alone gives no gain at the same number of steps.
On real Eyeful Tower photos, default guidance brings the prior back: retention reaches only $\beta = 0.40$, and forward
retention drops from $0.85$ to $0.73$.
We then continue training with DiffusionNFT~\cite{zheng2025diffusionnft} on a Pi3 reward that penalizes errors in
height, pitch, and yaw as well as low-confidence pixels (supplementary material).
The held-out reward improves from $-3.44$ to $-2.96$ over 15 epochs, but height retention improves only slightly on
synthetic starts and not measurably on real photos.

\begin{figure}[t]
  \centering
  \includegraphics[width=\linewidth]{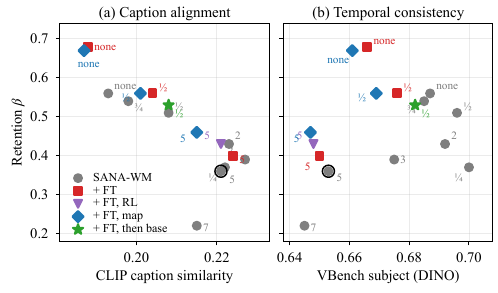}
  \caption{\textbf{Height retention against caption alignment and temporal consistency in SANA-WM} (real starts, 52
    turns per setting).
    Gray: the base model under guidance scales 2--7, no guidance, or the first \nicefrac14, \nicefrac12, or
    \nicefrac34 of the 60 steps unguided (circled: default).
    Colored: the fine-tuned models of \cref{tab:remedies} (star: late steps run by the base model).}
  \label{fig:tradeoff}
\end{figure}

\paragraph{Unguided fine-tuning keeps the most height at a cost in quality.}
Without text guidance, fine-tuning gives the highest real-start retention of all settings (\cref{fig:remedies}a,b),
with $\beta = 0.68$, 79\% of clips kept, and $\times1.12$ on the lowest tier.
The floor-map variant is close behind ($\beta = 0.67$).
On synthetic starts the two reach $\beta = 0.56$ and $0.59$, but their rollouts still drift toward default heights of
\SI{0.90}{m} and \SI{0.94}{m}.
The cost is perceptual quality: real-start MUSIQ drops by $5.3$ with plain fine-tuning and by $1.8$ with the floor map
(\cref{fig:remedies}c).
Without guidance the fine-tuned models also match the caption less well, while subject consistency stays above the
default (\cref{fig:tradeoff}).
Adding guidance in the second half of denoising lowers retention to $\beta = 0.56$ and does not restore quality (MUSIQ
$43.2$ against $47.6$ for the base model).
Running these guided late steps with the base model gives almost the same retention and quality, so the quality
loss is already fixed in the unguided first half, where camera height is also set (\cref{sec:mechanism}).

\section{Conclusion}
\label{sec:conclusion}

The six open-weight interactive video world models we evaluate do not keep the camera height of their starting frame.
On \bench, all six pull both floor-level and overhead viewpoints toward human eye level during
rotational outpainting, and this drift alone shifts the trajectories predicted by visual navigation policies.
In Matrix-Game, the resulting rollouts are geometrically split: local floor motion stays consistent with constant
elevation while newly outpainted room structure implies a higher camera.
We trace this pull to human-eye-level priors in large-scale video corpora: in SANA-WM, neither retaining the start
frame in context nor supplying explicit textual or dense geometric height inputs prevents it, whereas early-step
classifier-free text guidance strengthens it.

\paragraph{Limitations and outlook.}
Because generated rollouts lack ground-truth 3D geometry, our primary metric relies on multi-view floor reconstruction
and requires visible ground surfaces, which we cross-validate against VGGT, monocular metric depth, and learning-free
floor homographies.
The navigation analysis measures shifts in predicted paths in open loop, and closed-loop effects remain untested.
Our training and guidance interventions also center on SANA-WM, and we leave remedies for causal streaming models to
future work.
Finally, when we combine fine-tuning with unguided sampling, the quality loss arises in the same early denoising steps
that set camera height, so this post-hoc fix trades one for the other.
Height-diverse pretraining data and guidance that does not strengthen the eye-level prior may avoid this trade-off, and
world-model benchmarks should report camera-height retention alongside trajectory adherence.

{
    \small
    \bibliographystyle{ieeenat_fullname}
    \bibliography{main}
}

\end{document}